\documentclass[10pt,letterpaper]{article}

\usepackage[utf8]{inputenc}
\usepackage{url}
\usepackage{epsfig}
\usepackage{graphicx}
\usepackage{subfigure}
\usepackage{mathtools}
\usepackage{algpseudocode}
\usepackage{algorithm}
\usepackage{amsmath}
\usepackage{relsize}
\usepackage{mathrsfs}
\usepackage{bm}
\usepackage{amsfonts}
\usepackage{color}
\usepackage{xcolor}
\usepackage{verbatim}
\usepackage{subfig}

\usepackage[right]{lineno}

\usepackage[utf8]{inputenc} 
\usepackage{url}            
\usepackage{booktabs}       
\usepackage{amsfonts}       
\usepackage{nicefrac}       
\usepackage{microtype}      
\usepackage{lipsum}
\usepackage{fonttable}
\usepackage{graphicx}
\usepackage{algpseudocode}
\usepackage{algorithm}
\usepackage{everypage}
\usepackage{xspace}
\usepackage{fancyhdr} 
\usepackage{changepage}
\usepackage{multicol}
\usepackage{lipsum}       
\usepackage{lastpage}       
\usepackage{booktabs}
\usepackage{amsmath}
\usepackage{relsize}
\usepackage{subfigure}
\usepackage{rotating}
\usepackage{url}
\usepackage{bm}

\newcommand{\bx}{\mathbf{x}}

\newcommand{\bY}{\bm{Y}}

\newcommand{\btheta}{\bm{\theta}}
\newcommand{\bgamma}{\bm{\gamma}}
\newcommand{\bbeta}{\bm{\beta}}

\newcommand{\sref}[1]{Sec.~\ref{sec:#1}}
\newcommand{\qref}[1]{Eq.~(\ref{eqn:#1})}

\newcommand{\fref}[1]{Fig.~\ref{fig:#1}}

\definecolor{Gray}{gray}{.25}

\begin{document}
\vspace*{0.1in}

\begin{center}

{\Large
\begin{center}
\textbf{Towards Scalable Gaussian Process Modeling}
\end{center}
}

\bigskip
Piyush Pandita\textsuperscript{*},
Jesper Kristensen\textsuperscript{},
Liping Wang\textsuperscript{}
\\
\bigskip
Probabilistics and Optimization Team, GE Research, Niskayuna, New York, 12309
\\
\bigskip
* piyush.pandita@ge.com

\end{center}

\begin{abstract}
Numerous engineering problems of interest to the industry are often characterized by 
expensive black-box objective function evaluations. These objective functions could be physical experiments or computer simulations. Obtaining a comprehensive idea of the problem and/or performing subsequent optimizations generally requires hundreds of thousands of evaluations of the objective function which is most often a practically unachievable task. Gaussian Process (GP) surrogate modeling replaces the expensive function with a cheap-to-evaluate 
data-driven probabilistic model. While the GP does not assume a functional form of the problem, it is defined by a set of parameters, called hyperparameters, that need to be learned from the data.
The hyperparameters define the characteristics of the objective function, such as smoothness, magnitude, periodicity, etc. Accurately estimating these hyperparameters is a key ingredient in developing a reliable and generalizable surrogate model. Markov chain Monte Carlo (MCMC) is a ubiquitously used Bayesian method to estimate these hyperparameters. At GE’s Global Research Center, a customized industry-strength Bayesian hybrid modeling framework utilizing the GP, called GEBHM, has been employed and validated over many years. GEBHM is very effective on problems of small and medium size, typically less than 1000 training points.
However, the GP does not scale well in time with a growing dataset and problem dimensionality which can be a major impediment in such problems. For some challenging industry applications, the predictive capability of the GP is required but each second during the training of the GP costs thousands of dollars.  In this work, we apply a scalable MCMC-based methodology enabling the modeling of large-scale industry problems. Towards this, we extend and implement in GEBHM an Adaptive Sequential Monte Carlo (ASMC) methodology for training the GP. This implementation saves computational time (especially for large-scale problems) while not sacrificing predictability over the current MCMC implementation. We demonstrate the effectiveness and accuracy of GEBHM with ASMC on four mathematical problems and on two challenging industry applications of varying complexity.\\

\end{abstract}

\section{Introduction}
\label{sec:intro}
Computer simulators \cite{schonlau1997computer} and/or sophisticated in-house laboratory experiments \cite{flournoy1993clinical} representing physical phenomena generally pose the problem of being computationally or logistically challenging. In scenarios where extracting information \cite{jaynes1957information,kullback1997information} about an underlying physical process to forecast or predict its behaviour is critical, generating hundreds of thousands of experimental data points becomes infeasible. To alleviate the above issue, surrogate modeling (also called predictive modeling) is a popular approach towards obtaining an inexpensive representation of the simulation or experiment. 

Among the myriad surrogate modeling \cite{gunst1996response} techniques, Gaussian Process (GP) \cite{williams2006gaussian} regression (GPR) is a well established technique in the area of probabilistic modeling of expensive functions. GPR is a non-parametric machine learning technique in the sense that it does not make assumptions about the functional form of the function being modeled. However, the GP model does have parameters, called hyperparameters, that capture characteristics such as periodicity, smoothness, measurement noise, etc., of the underlying function. These hyperparameters are learned from the data (limited set of observations) using empirical Bayesian techniques like maximum likelihood estimation
\cite{pan2002maximum}, or fully Bayesian \cite{kennedy2001bayesian} methods based on Markov chain Monte Carlo (MCMC) \cite{robert2013monte,gelman1995bayesian}. For a typical GP the number of hyperparameters is on the order of 10-100.
Furthermore, the predictive accuracy of the GP is highly sensitive to the values of the hyperparameters thus requiring sophisticated optimization. The high dimensionality of the hyperparameter space prompts a rigorous MCMC treatment \cite{cowles1996markov} to learn the hyperparameters. Also, the Bayesian MCMC \cite{gilks1995markov,carlo2004markov,green1995reversible} methods allow quantification of uncertainties \cite{ghahramani2003bayesian} around the estimates of the hyperparameters, which makes possible drawing samples from the space of the underlying function.

Recent work on extending GPs to big-data applications has focused on deriving variational representations of GPs~\cite{peng2017asynchronous}, constructing sparse approximations of GPs~\cite{hensman2013gaussian,quinonero2005unifying} and training local GPs using informative subsets of the data~\cite{snelson2007local,lee2017hierarchically}.
However, for most of these methods, the modeling assumes the parameters of the covariance kernel fixed to point estimates that are obtain using optimization or are considered known.
A recent review of methods for extending GPs for big-data applications can be found in~\cite{liu2018gaussian}.
In this paper, we restrict our focus to Bayesian GPs where the covariance parameters are inferred using MCMC.
 
GE's Global Research Center has its own implementation of fully Bayesian modeling, called GE Bayesian Hybrid Modeling (GEBHM), which has been applied \cite{ghosh2018bayesian,kristensen2016expected} successfully to various engineering problems over many years.
Given the greater predictability achieved by using MCMC, the  MCMC method comes with its own pitfalls \cite{o1987monte}. 
For example, in problems involving huge training data sets, on the order of $N=1000$ data points, the excessive computational training time is a direct consequence of a large number of GP covariance matrix inversion operations, scaling as $O(N^{3})$. The work done in \cite{leithead2007n} has focused on circumventing this problem by obtaining approximations to the inverse of the matrix. Scalability of the MCMC methods with varying complexity of the problem, such as increasing number of input dimensions \cite{haario2004markov}, multiple correlated functions, etc., can break down when applied to challenging industrial problems.
 
 Sequential Monte Carlo (SMC) \cite{doucet2001introduction,diaconis2003sequential} methods offer a promising alternative considering the availability of multiple individual processor elements (PEs), also called cores, and high performance computing environments. The inherent parallelizability paves the way for SMC methods to be used in fitting the hyperparameters through a form of importance sampling and reduce the required number of matrix inversions per PE. Moreover, SMC methods can be tweaked by the user based on the type of problem through multiple \emph{knobs} which determine characteristics of the SMC such as number of simultaneously running MCMC chains, number of MCMC steps per chain, discretization of the posterior distribution into finite samples, etc. This magnanimity is an innate feature of most SMC algorithms \cite{andrieu2001sequential,andrieu2010particle}, the theory of which guarantees convergence under specific assumptions \cite{bilionis2012free,bilionis2015crop}. We apply an adaptive SMC (ASMC) method \cite{kristensen2013relative}, by leveraging \emph{pySMC} (a package written in the PYTHON programming language). The details of the pySMC library are provided at \url{https://github.com/
ebilionis/pysmc}. The ASMC method samples according to the same kernel as GEBHM's standard MCMC, albeit reducing the number of covariance matrix inverse computations by an order of magnitude. This difference is critical to the ASMC alleviating the burden of excessive computational time in GP model training. We extend the capabilities and flexibility of the ASMC by allowing for a predefined set of importance sampling distributions which could be provided by the user, based on the complexity of the problem. As an addendum to the methodology, fine-tuning of the widths of proposal distributions of the hyperparameters is done on the fly, independently.

The purpose of this study is to verify the scalability of ASMC, its higher predictive accuracy, and to demonstrate the savings in time (which translate to cost savings) that can be leveraged henceforth. The investigation is carried out using six problems with varying types of complexity in terms of the input dimensionality, the number of outputs, the size of the training data, etc. The comparison tests are done using multiple PEs on a workstation computer to highlight the ease of implementation and usability of the ASMC. Further comparison tests include the results obtained by the seamless scalability achieved by ASMC using high performance computing (HPC) clusters. Out of the six chosen problems four are synthetic in nature with different dimensionality and training data size and two stem from GE industrial applications.

The outline of the paper is as follows. We start \sref{metho} by providing some basic mathematical definitions of GPR and the details of the posterior of the
hyperparameters of the GP. Our numerical results are presented in \sref{results}. The impact of the ASMC on problems with large training data sets is highlighted in \sref{scalability}.
In particular, in \sref{toy_small}, \sref{toy_medium} and \sref{toy_large}, we validate our approach using three synthetic
problems with varying input dimensionality.
In \sref{flame_transfer} and \sref{combustions}, we apply the ASMC methodology to solve a challenging steam turbine compressor problem and a multi-objective combustion
problem, respectively. We present our conclusions in \sref{conc}.

\section{Methodology}
\label{sec:metho}
\subsection{Gaussian process regression}
\label{sec:gp}
The surrogate models that GEBHM builds for the overall objective(s) for the problem are data-driven statistical models that rely on the Bayesian nature of GPR \cite{williams2006gaussian}.
GPs are non-parametric meaning, they do not assume a model form (think: polynomial of degree $n$, among others). GPs try to map inputs to outputs by superposing several Gaussian distribution functions. GPs can accurately model complex variations in the data, they can capture discontinuities effectively and can produce very accurate surrogate statistical models with comparatively less data. An important feature is that GPs offer a measure of the quality of the surrogate statistical model and direct us to areas of input space where we need more data, for example \cite{kristensen2017adaptive}. We now go over the finer details of GPR where we define the posterior distribution of the hyperparameters which we want to sample from.
We use a zero mean for the GP. The covariance of the GP is computed based on a kernel that is a function of the distance between the inputs. For a multi-objective problem the covariance matrix is a block diagonal matrix comprising of the individual covariance matrices for each of the objectives. The covariance for an $m$ output, $d$ input, problem with $N$ training data points is shown below: 
\begin{equation}
\Sigma=
\begin{bmatrix}
\Sigma^{(1)} &		&		\\
				& \ddots & 	\\
				&		&\Sigma^{(m)}
\end{bmatrix},
\label{eqn:cov_overall}
\end{equation}
where each of the individual elements along the diagonal represent the covariance matrix of dimension $N \times N$ of a single output. 
These can be more generally expressed as in \qref{cov_ind}.
\begin{equation}
\Sigma^{(k)}_{ij} = \frac{1}{\lambda_{z}^{(k)}}\sum_{l=1}^{d} \exp\left[-\beta_{l}^{(k)}{(x_{i,l}-x_{j,l})^2}\right] + I\frac{1}{\lambda_{s}^{(k)}} + I\frac{1}{\lambda_{o}} 
\label{eqn:cov_ind}
\end{equation}
where $i,j \in \{1,\cdots, N\}$, $ k \in \{1, \cdots, m\}$, $l \in \{1, \cdots, d\}$, $x_{i,l}$ is the value of the $i$th training point in the $l$th input dimension, and let $\bbeta^{(k)}=\{\beta_{1}^{(k)} \cdots \beta_{d}^{(k)}\}$.
The hyperparameters of the GP model are $\btheta = \{\bbeta^{(k)}, \lambda_{z}^{(k)}, \lambda_{s}^{(k)},  \lambda_{o}\}_{k=1}^{m}$. 
The $\bbeta^{(k)}$ parameters represent the inverted length-scales of the covariance kernel. $\lambda_{z}^{(k)} $ is also called the signal-strength which is proportional to the scale of the magnitude for the objective space and $\lambda_{s}^{(k)}$ represents the precision of the random Gaussian measurement noise. $\lambda_{o}$ represents the precision of the random Gaussian measurement noise to account for the overall randomness between the multiple outputs. It can be inferred from above that for our problems of learning the model's hyperparameters, it is always a problem of estimating $m\times (d+2) + 1$ number of parameters. 
The likelihood of observing the training data for a selected set $\btheta$ of hyperparameters is:
\begin{equation}
L(\bY|\btheta) = \frac{1}{|\Sigma|^{\frac{1}{2}}}\exp(-\frac{1}{2}\bY^{T}\Sigma^{-1}\bY),
\label{eqn:likelihood}
\end{equation}
where the $i, j$th element of the matrix $\bY$ is the $j$th output value for the $i$th training data point.
The conditional posterior of the hyperparameter set $\btheta$ can be written as follows:
\begin{equation}
p(\btheta|\bY) \propto L(\bY|\btheta)p(\lambda_{o})\prod_{k=1}^{m}p(\bbeta^{(k)})p(\lambda_{z}^{(k)})p(\lambda_{s}^{(k)}).
\label{eqn:posterior}
\end{equation}
The posterior given in \qref{posterior} is the so-called \emph{target distribution} known only up to a proportionality constant. The sequence of sampling distributions in the ASMC are derived from the functional form given in \qref{posterior}.
\subsection{MCMC in GEBHM}
\label{sec:bhm_mcmc}
The implementation of MCMC in GEBHM works in two phases: a) the initialization phase and b) the main chain.  
The initialization phase of the MCMC solves the purpose of selecting the proposal width of the proposal distributions for the main (full) chain of the MCMC. This initialization process can only be performed on a single PE, thus limiting the computational power offered by multiple cores. The main chain can use multiple cores, but the single core operation of initialization does contribute to a significant percentage of the total cost of the MCMC in GEBHM. The complete details of the initialization process have been omitted to protect company proprietary information. The MCMC algorithm uses a Metropolis algorithm \cite{gelman1996efficient} for making jumps (steps) during the main chain.
\subsection{Adaptive sequential Monte Carlo}
\label{sec:asmc_math}
The sequential Monte Carlo methodology (also known as particle filtering in some literature) \cite{gordon1993novel} essentially uses a coherent sequence of distributions to do MCMC sampling from. This sequence of distributions starts with a distribution mimicking a uniform distribution and progressively moves closer (in analytical form) to the posterior distribution \qref{posterior} (in our case).
A sequence of $n$ sampling distributions based on the empirical form given in \qref{pi_gamma} can be built with $\gamma_{0} \sim 0 $ and $\gamma_{n} = 1$. Thus, the sampling process can be understood as a slow progression from a non-informative \emph{uniform} distribution to the target posterior distribution  \qref{posterior}. The jumps between states for all sampling distributions are based on a Metropolis stepper (same as the one being used for MCMC in GEBHM).
The conditional posterior in \qref{posterior} is reproduced in \qref{smc_true} while we shun the dependence on $\bY$ to clearly represent the form of the sampling distributions of the ASMC as a function of only the hyperparameters $\btheta$.
The sampling distributions are illustrated in \qref{smc_dist}. Each $p_{i}(\btheta)$  is a sampling distribution and each chain (ASMC particle) takes $num_{asmc}$ steps before moving on to a different value of $\gamma$. 
\begin{eqnarray}
p(\btheta)&\propto&\pi(\btheta)
\label{eqn:smc_true}\\
{p_{i}(\btheta)}&\propto&{\pi_{i}(\btheta)}
\label{eqn:smc_dist}\\
{\pi_{i} (\btheta)}&=&{\pi(\btheta)}^{\gamma_{i}}.
\label{eqn:pi_gamma}
\end{eqnarray}
At each $\gamma_{i}$ the corresponding distribution ${p_{i}(\btheta)}$ is described as a discrete particle approximation as shown in \qref{particle_approx}.
\begin{equation}
{p_{i}(\btheta)} \approx {\mathlarger{\sum}}_{j=1}^{N}w^{j_{i}}\delta(\btheta-\btheta^{j_{i}})
\label{eqn:particle_approx}
\end{equation}
These finite number of samples are called particles (parallel MCMC chains). At each value of $\gamma$ each particle takes a predefined number of steps $num_{asmc}$. Thus, each particle evolves its own MCMC chain based on the Metropolis stepper.
The values of $\gamma$ visited by the algorithm are governed by a statistic known as the \emph{effective sample size} (\emph{ess}). The \emph{ess} is calculated as defined in \qref{ess}. Thus, when all particles have equal weights the \emph{ess} is equal to the number of particles, $N$.  The \emph{ess} guides the methodology finding the next $\gamma$ i.e. $\gamma_{i+1}$  as shown in \qref{ess}.
\begin{eqnarray}
ess|\gamma_{i}&=&{\mathlarger{\sum}}_{j=1}^{N}\frac{1}{{w^{j_{t}}}^{2}}
\label{eqn:ess}\\
ess|\gamma_{i+1}&=&ess\_reduction \times ess|\gamma_{t},
\label{eqn:ess_reduction}
\end{eqnarray}
where $ess_{reduction}$ is a predefined threshold $\in{(0, 1)}$.
The $\gamma_{i+1}$ is usually selected via an algorithm like \cite{brent1980improved}.
For more details on the \emph{ess} and the selection of $\gamma_{i+1}$ the reader is referred to the work of \cite{bilionis2015crop,bilionis2012free}.
The dynamic selection of the set of $\gamma$s for a single ASMC requires special attention in the industrial context.
For some problems the dynamic selection process can be too time consuming. 
Towards ameliorating this we replace the dynamic selection of the $\gamma$ parameter with a prespecified set of $\gamma$s called the \emph{grid}. The \emph{grid} used in this works allows uniform sized jumps for the $\gamma$ parameter but we expect better results with a non-uniform \emph{grid}.
Consider \qref{ess_reduction}, the number of $\gamma$s visited by the ASMC can vary based on factors such as the size of training data, number of input dimensions, etc. The \emph{grid} thus constrains the freedom of the algorithm especially when time is a greater priority for a user than small gains in accuracy. 
We list the steps taken by GEBHM's ASMC methodology in algorithm \ref{alg:asmc}.
\begin{algorithm}[htb]
\caption{ASMC in GEBHM.}
\begin{algorithmic}[1]
\Require
		   number of steps per particle $num_{asmc}$,
		   \emph{grid}: $\bgamma$,
		   MCMC stepper, 
	 \State Initialize the particles (chains) via MCMC stepper.
            \For {$i \in [1, n]$}
                \State Do $num_{asmc}$ steps for each particle.
                \State \emph{tune} the proposal widths of the proposal distributions for each hyperparameter based on the			respective acceptance ratios from the MCMC stepper.
                \State $\gamma \leftarrow \bgamma_{i}$
    			\EndFor
\end{algorithmic}
\label{alg:asmc}
\end{algorithm}
The added advantage of the ASMC is the lack of need to pre-compute the proposal widths (as is the case for MCMC in GEBHM) as it adaptively scales the proposal widths on the fly for each parameter. The exploration of posteriors of parameters with multiple modes is a problem that the ASMC is semantically designed to tackle.
\section{Results}
\label{sec:results}
We first present our numerical results demonstrating the ASMC's ability to scale well with large training data sets.
\subsection{Toy problem 1: Scalability}
\label{sec:scalability}
As mentioned in \sref{intro}, the standard MCMC implementation makes the GP model training process cumbersome. In cases with large training data, the aforementioned problem worsens with the whole data set not being put to use for training the GP model. Reducing the frequency of the computationally challenging part is critical to the consequence of applying a MCMC method to GP model training. We apply the ASMC to the following 10 dimensional non-linear mathematical problem:
\begin{eqnarray}
y(\bx)&=&3\sin(x_{1})x_{2} + \cos(x_{3})\sin(x_{4}) + \sin(x_{5})\sin(x_{6}) \nonumber \\
&& + \sin(x_{7}) + \sin(x_{8}) + 7x_{9} + 6x_{10}
\label{eqn:scalability}
\end{eqnarray}
\qref{scalability} enables us to generate a large training data set (greater than 1000 points).
The performance of the two algorithms, GEBHM and ASMC, is highlighted in \fref{scalability} (a),  (b) and (c). ASMC's ability to 
handle large sets of data is clearly visible in \fref{scalability} (a) where the standard MCMC takes almost four times as long as the ASMC with around 1000 training data points. Secondly, the ASMC's GP model building times are comparatively low even with training data size greater than 3000 points. This is accomplished due to ASMC's seamless operational scalability to HPC environments. \fref{scalability} (b) highlights the predictive accuracy achieved by ASMC (which is mainly a satisficing objective) being equal to or better than the standard MCMC.
The MCMC settings were kept at 6800 steps including 2000 steps for the initialization phase.
The savings in absolute time shown in \fref{scalability}(c) provide an idea of the potential improvement made by using ASMC for GP model building with high training data problems. 
It is important to consider the non-parallelizability of the initialization (shown in purple) of the MCMC, which limits the depth of impact that can be achieved by using a larger number of cores (greater than 12) for the GEBHM. The times represented by the initialization will remain fixed irrespective of the number of cores, thus making the comparison with ASMC even more realistic. 
\begin{figure}[htp]
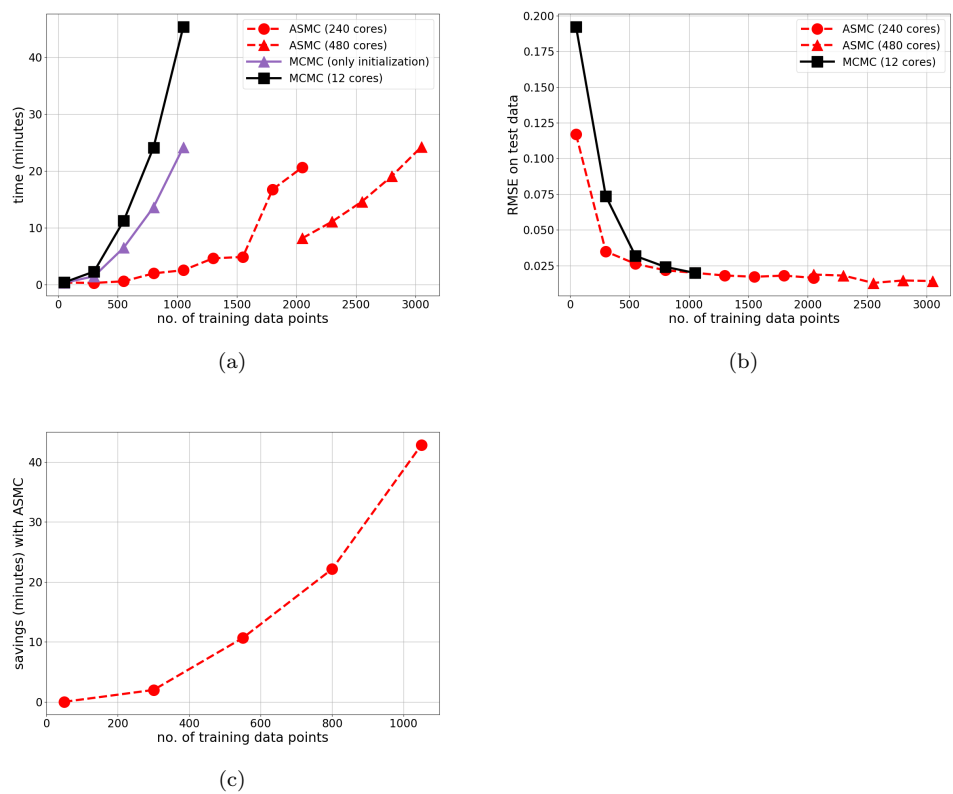

\subfigure[]{\includegraphics[width=0.5\textwidth]{{{time_data_scalability}.png}}}
\subfigure[]{\includegraphics[width=0.5\textwidth]{{{rmse_data_scalability}.png}}}
\subfigure[]{\includegraphics[width=0.5\textwidth]{{{time_savings_data_scalability}.png}}}
\caption{Subfigures~ (a), (b) and (c) show the scalability, predictive accuracy and savings in time for the ASMC compared to MCMC for a large training data set problem respectively.  The number of particles for the ASMC is  480, for all runs.}
\label{fig:scalability}
\end{figure}
For the scalability problem we allow the ASMC to identify the set of sampling distributions on the fly, see \qref{ess_reduction}, hence not specifying the \emph{grid}.
\subsection{ASMC and MCMC settings}
Now, we move on to the comparison of the ASMC with GEBHM's MCMC on three synthetic problems. We plot the RMSE for the ASMC obtained using the final particle approximation for different number of particles, keeping the number of cores on a workstation fixed. As the number of particles increases, the time taken to build the GP model also increases and the corresponding RMSE shows gradual decrease. This is done because plotting the RMSE by using particles from the different sampling distributions of a single ASMC run would be misleading simply because the sampling distributions visited by the ASMC are different from one another. This is a crucial step, where the ASMC differs from the conventional MCMC (where the sampling distribution remains the same i.e. the posterior distribution of the hyperparameters known up to a proportionality constant \qref{posterior}). \\
For workstation computations the ASMC settings are fixed to 6 cores, and the number of particles at 6, 12, 30, and 60. For HPC computations we vary the number of cores and fix the number of particles equal to the number of cores for each run of the ASMC. The number of cores for the HPC runs of the ASMC for all problems are 24, 48, 120, 240, and 480. For ASMCs performance on the HPC with greater than 480 particles (for the toy problem in \sref{toy_large}), the number of cores is fixed at 480 (the highest number available during our experiments on the HPC).

However, for the standard MCMC, the RMSE on the test data is plotted against time as the chains evolve. The settings for the standard MCMC are 5800 steps including 1000 steps for initialization. The $\gamma_{0}$ values are 0.001 and 0.00001 for the ASMC's runs on the workstation and the HPC, respectively.
\subsection{Toy problem 2}
\label{sec:toy_small}
The four dimensional problem, is a single-objective problem with a known analytic form of the objective. The objective function is a quadratic function of the four input variables. For safeguarding proprietary information of GE's business partners, the full functional form of the input-output relation has been omitted from the text.\\
\begin{figure}[h]
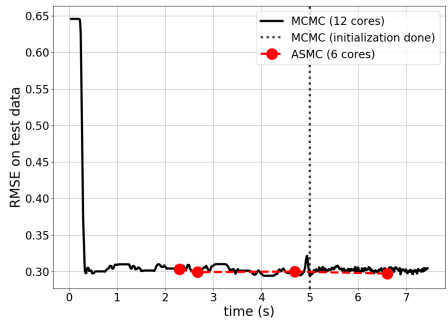

\centering
{\includegraphics[width=0.5\textwidth]{{{rmse_time_small_pc}.png}}}
\caption{Root mean squared error versus time taken to build the GP model for the two algorithms for the four dimensional toy problem. Number of particles (red dots) are 6, 12, 30, and 60, across the time axis for ASMC.}
\label{fig:toy_small}
\end{figure}
The purpose of comparing the two algorithms on a low-dimensional problem is to demonstrate the ability of the ASMC to replace the standard MCMC on a workstation, before showing comparisons between the two methodologies by leveraging HPC platforms.
\fref{toy_small} shows the savings in time on a workstation with the ASMC using 6 cores and the GEBHM standard MCMC using 12 cores after the completion of the initialization (done on a single core). Each red dot in \fref{toy_small} represents the RMSE on test data after the completion of the ASMC GP model building, as the number of particles for the ASMC is varied across time.
For the standard MCMC, the RMSE on test data during initialization is separated from the rest of the particles by the vertical dotted line. The standard MCMC RMSE versus time is plotted by computing the RMSE at each time step of the MCMC chain. 
Higher dimensional problems are used to compare the two algorithms with ASMC scaled-up on the HPC.
The number of steps per ASMC particle is equal to one. The \emph{grid} is fixed at 10 steps and 20 steps for the workstation and HPC, respectively. 
\subsection{Toy problem 3}
\label{sec:toy_medium}
We use an 18 dimensional, single-objective problem to further demonstrate the effectiveness of the methodology. The problem is a torsion vibration problem where the high natural frequency of a three-shaft and two-disk system is the objective function. The input variables include the diameter, length, rigidity and weight density of the three shafts and the diameter, thickness, weight density of the two disks. A diagram representing the structure of the system is shown in \fref{toy_medium_pic}.
The problem can be mathematically represented as follows:
\begin{eqnarray}
y&=&\frac{\sqrt{\frac{-b + \sqrt{b^{2} -4ac}}{2a}}}{2\pi}  \nonumber \\
a&=&1 \nonumber \\
b&=&-\Bigg(\frac{K_{1} + K_{2}}{J_{1}} + \frac{K_{2} + K_{3}}{J_{2}}\Bigg) \nonumber \\
c&=&\frac{K_{1}K_{2} + K_{2}K_{3}+K_{3}K_{1} }{J_{1}J_{2}} \nonumber 
\label{eqn:toy_medium}
\end{eqnarray}
where, 
\begin{eqnarray}
K_{i}&=&\frac{\pi G_{i}d_{i}}{32L_{i}} \nonumber \\
M_{j}&=&\frac{\pi t_{j}\rho_{j}D_{j}}{4g}  \nonumber \\
J_{j}&=&\frac{1}{2}M_{j}\Bigg(\frac{D_{j}}{2}\Bigg)^{2}
\label{eqn:toy_medium_params}
\end{eqnarray}
for $i \in \{1, 2, 3\}$ and $j \in \{1, 2\}$. The $d_{i}s$, $L_{i}s$, $G_{i}s$ and $\lambda_{i}s$ represent the diameter, length, rigidity and the weight density of the three shafts and $D_{j}s$, $t_{j}s$ and $\rho_{j}s$ represent the diameter, thickness, and weight density of the two disks, respectively.\\
The convergence analysis of the two algorithms is shown in \fref{toy_medium}.
The number of steps per ASMC particle is five. The \emph{grid} is fixed at 10 steps and 20 steps for the workstation and HPC, respectively. 
\begin{figure}[h]
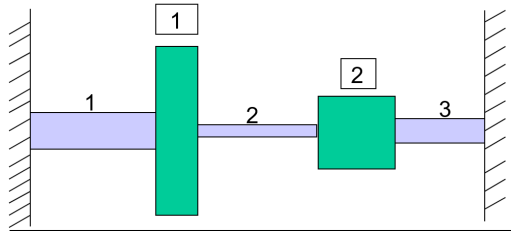

\centering
{\includegraphics[width=0.5\textwidth]{{{toy_medium_pic}.png}}}
\caption{The setup of the torsion vibration problem.}
\label{fig:toy_medium_pic}
\end{figure}
\begin{figure}[h]
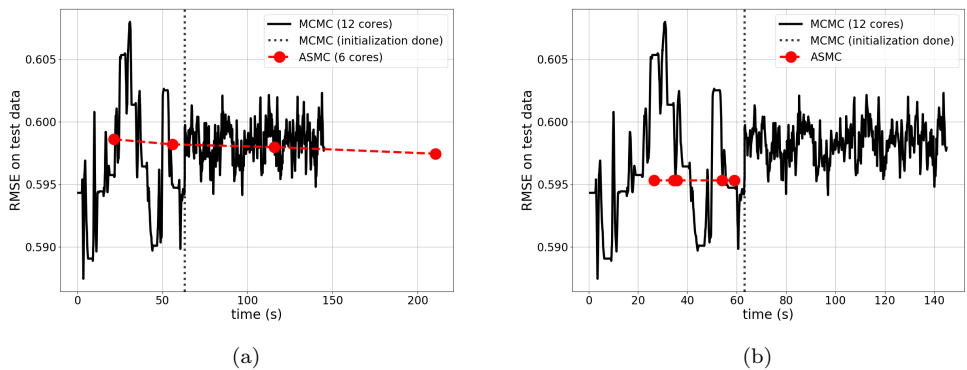

\subfigure[]{\includegraphics[width=0.5\textwidth]{{{rmse_time_medium_pc}.png}}}
\subfigure[]{\includegraphics[width=0.5\textwidth]{{{rmse_time_medium_hpc}.png}}}
\caption{Subfigure~(a) Number of particles on a workstation (red dots) are 6, 12, 30, and 60, and Subfigure~(b) number of particles on the HPC (red dots) are 24, 48, 120, 240, and 480, across the time axis, respectively for the torsion problem.}
\label{fig:toy_medium}
\end{figure}
\subsection{Toy problem 4}
\label{sec:toy_large}
A 100 dimensional non-linear single-objective problem poses a strong challenge in terms of the number of input dimensions to test the performance of both methodologies. The explicit details of the functional relationship have been omitted from the manuscript.\\
The convergence is terms of the RMSE on test data as a function of time is shown in \fref{toy_large}. The convergence for the workstation based runs can be seen in \fref{toy_large} (a), where the ASMC reaches almost similar accuracy as the standard MCMC in less than half the time (with 30 particles). Similar convergence trends can be seen for the ASMCs runs on the HPC with varying number of particles (24, 48, 240, 480, 960 and 2400) with time.
The number of steps taken by each ASMC particle is one for both (workstation and HPC) cases. The \emph{grid} is fixed at 10 steps and 20 steps for the workstation and HPC, respectively. The finer \emph{grid} considered for the HPC is to gain greater predictive accuracy while relying on the HPC (large number of cores) to compensate for the computational overheads incurred.
\begin{figure}[h]
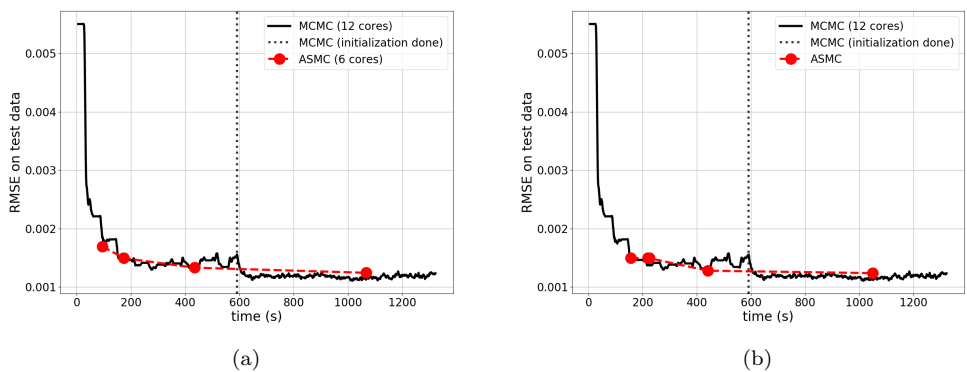

\subfigure[]{\includegraphics[width=0.5\textwidth]{{{rmse_time_large_pc}.png}}}
\subfigure[]{\includegraphics[width=0.5\textwidth]{{{rmse_time_large_hpc}.png}}}
\caption{Subfigure~(a) Number of particles on a workstation (red dots) are 6, 12, 30, and 60, and Subfigure~(b) number of particles on the HPC (red dots) are 48, 240, 480, 960 and 2400, across the time axis, respectively for the 100 dimensional problem.}
\label{fig:toy_large}
\end{figure}
\section{Industry problems}
\subsection{Steam turbine compressor problem}
\label{sec:flame_transfer}
This is a five dimensional, single-objective design optimization problem, where the training data set is observed from temperature experiments. The goal is to obtain the best system design given requirements on the temperature distributions.
The number of steps taken by the ASMC particles during the workstation based runs is five and for the runs on the HPC is one. The \emph{grid} is fixed at 10 steps for both, workstation and HPC.
\begin{figure}[h]
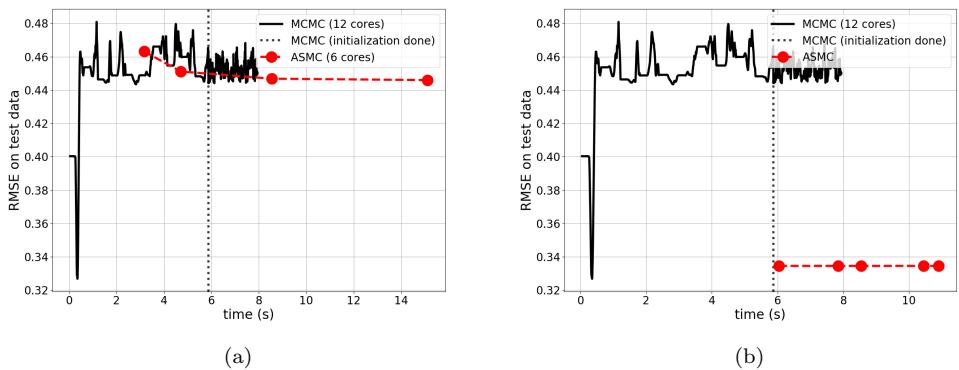

\subfigure[]{\includegraphics[width=0.5\textwidth]{{{rmse_time_flame_transfer_pc}.png}}}
\subfigure[]{\includegraphics[width=0.5\textwidth]{{{rmse_time_flame_transfer_hpc}.png}}}
\caption{Subfigure~(a) Number of particles on a workstation (red dots) are 6, 12, 30, and 60, and Subfigure~(b) number of particles on the HPC (red dots) are 24, 48, 120, 240, and 480, across the time axis, respectively for the steam turbine compressor problem.}
\label{fig:flame_transfer}
\end{figure}
The savings in time and, for this case noticeable, gains in predictive accuracy are shown in \fref{flame_transfer}. Not only does the ASMC show explicit time savings, but with the HPC  it also results in greater predictive accuracy than the GEBHM standard MCMC as shown in \fref{flame_transfer} (b). This indicates the dense representation of the sampling distributions made possible by using the HPC (where each particle runs on a separate core).
\subsection{Combustion problem}
\label{sec:combustions}
The problem is a GE combustion test data problem, where we aim to build a GP model to predict two emission quantities as a function of three measured temperatures, three fuel flow parameters, and air flow. Thus, there are 9 hyperparameters that need to be estimated for each individual objective. The total number of hyperparameters for the GP model is 19. The MCMC settings for this problem are: 10000 steps with 1000 steps during the initialization.
\begin{figure}[h]
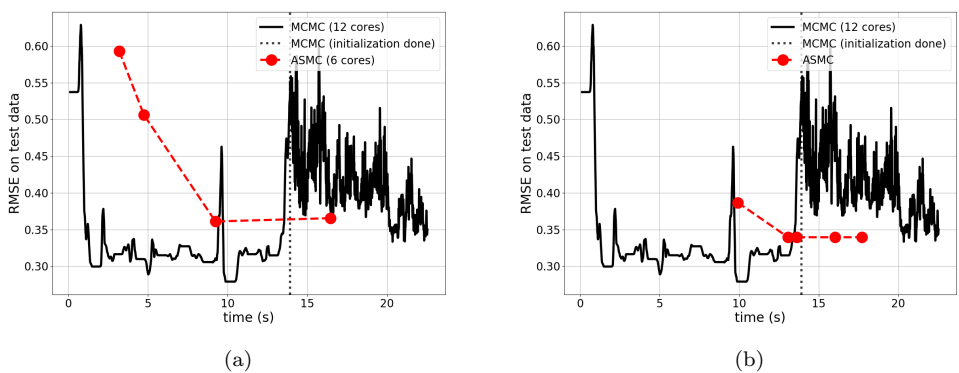

\subfigure[]{\includegraphics[width=0.5\textwidth]{{{rmse_time_combustions_y1_pc}.png}}}
\subfigure[]{\includegraphics[width=0.5\textwidth]{{{rmse_time_combustions_y1_hpc}.png}}}
\caption{Subfigure~(a) Number of particles on a workstation (red dots) are 6, 12, 30, and 60, and Subfigure~(b) number of particles on the HPC (red dots) are 24, 48, 120, 240, and 480, across the time axis, respectively for the first objective of the combustion problem.}
\label{fig:combustions_1}
\end{figure}
\begin{figure}[h]
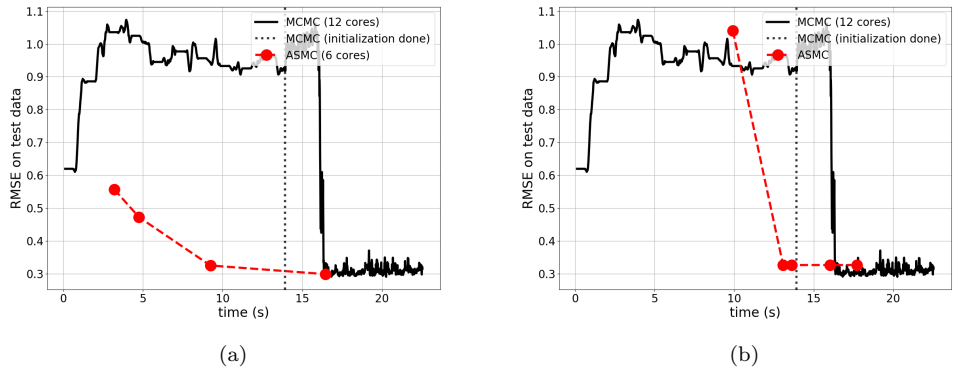

\subfigure[]{\includegraphics[width=0.5\textwidth]{{{rmse_time_combustions_y2_pc}.png}}}
\subfigure[]{\includegraphics[width=0.5\textwidth]{{{rmse_time_combustions_y2_hpc}.png}}}
\caption{Subfigure~(a) Number of particles on a workstation (red dots) are 6, 12, 30, and 60, and Subfigure~(b) number of particles on the HPC (red dots) are 24, 48, 120, 240, and 480, across the time axis, respectively for the second objective of the combustion problem.}
\label{fig:combustions_2}
\end{figure}
The significance of the results in \fref{combustions_1} and \fref{combustions_2} is noticeable from the perspective of the number of objectives. Not only does the scaled-up HPC ASMC do well in terms of computational time but it improves the predictive accuracy of the GP model compared to GEBHM's MCMC. The number of steps per ASMC particle is one for the workstation and five for the HPC runs. The \emph{grid} is fixed at 10 steps for both computing environments.

This provides a holistic overview of ASMC's performance on different types of problems. Problems of varying input dimensionality, varying number of objectives, and different training data sizes have been treated with the extended ASMC and compared to the current MCMC implementation. The ASMC does equally well or better compared to the MCMC in terms of predictive accuracy, while saving time by half or more in most of the challenging problems.
\section{Conclusions}
\label{sec:conc}
We demonstrate the working of the implementation of an ASMC method in GEBHM, and compare it to the performance of the existing semi-parallelizable MCMC algorithm on mathematical problems and challenging industrial problems with varying complexity. The main reason behind leveraging ASMC is the inherent flexibility and scalability offered by the algorithm's parallelizable nature, which is practically implementable with the availability of \emph{high-end} computational resources such as HPCs. The numerical results on toy problems of different input dimensionality show the versatility of the extended ASMC on both, workstation and HPC environments. This is augmented by extending the flexibility of the ASMC to allow the user to prespecify the sequence of sampling distributions. Thus, allowing the algorithm to balance savings in time and predictive accuracy. It is interesting to note that the ASMC has the inherent capability to treat different problems differently, but can also almost always perform well with a reasonable setting of the default parameters. The different parameters chosen with different problems in \sref{results} further highlights this nuance of the ASMC.
ASMC's performance on the combustion data problem, demonstrates its ability to perform equally well with multiple objectives.
Future directions of work on studying the algorithm include determining an optimal set of parameter settings for an arbitrary problem. This is important from the perspective of making the ASMC's implementation deliverable to users averse to the flexibility of the ASMC.
\section{Acknowledgements}
The authors wholeheartedly thank Dr. Ilias Bilionis, Purdue University for providing technical insights into this research.
\nolinenumbers
\bibliography{references}
\bibliographystyle{abbrv}
\end{document}